\documentclass[letterpaper, 10 pt, conference]{ieeeconf}  % Comment this line out if you need a4paper

\IEEEoverridecommandlockouts                              % This command is only needed if 
\usepackage{graphics} % for pdf, bitmapped graphics files
\usepackage{epsfig} % for postscript graphics files
\usepackage{mathptmx} % assumes new font selection scheme installed
\usepackage{times} % assumes new font selection scheme installed
\usepackage{amsmath} % assumes amsmath package installed
\usepackage{amssymb}  % assumes amsmath package installed
\usepackage{hyperref}
\usepackage{dblfloatfix} 

\usepackage{algorithm,algpseudocode}
\makeatletter
\newcommand{\algmargin}{\the\ALG@thistlm}
\makeatother
\newlength{\whilewidth}
\algdef{SE}[parWHILE]{parWhile}{EndparWhile}[1]
  {\parbox[t]{\dimexpr\linewidth-\algmargin}{%
     \hangindent\whilewidth\strut\algorithmicwhile\ #1\ \algorithmicdo\strut}}{\algorithmicend\ \algorithmicwhile}%
\algnewcommand{\parState}[1]{\State%
  \parbox[t]{\dimexpr\linewidth-\algmargin}{\strut #1\strut}}

\usepackage[draft]{pgf}
\usepackage{shapepar}
\usepackage{tikz}
\usetikzlibrary{shapes.geometric}

\title{\LARGE \bf
On the Design and Optimization of an Autonomous Microgravity Enabling Aerial Robot
}

\author{Juan-Pablo Afman$^{1}$, John Franklin$^{2}$, Mark L. Mote$^{3}$, Thomas Gurriet$^{4}$, Eric Feron$^{5}$*, % <-this % stops a space
\thanks{*This work was supported by the National Science Foundation}% <-this % stops a space
\thanks{$^{1}$J.P. Afman is with the Research Faculty of the Decision and Control Laboratory at the Georgia Institute of Technology, North Ave NW, Atlanta, GA 30332, USA
        {\tt afman@aerospace.gatech.edu}}%
        \thanks{$^{2}$J.K. Franklin is an Undergraduate Researcher of the Decision and Control Laboratory at the Georgia Institute of Technology, North Ave NW, Atlanta, GA 30332, USA
        {\tt jfranklin36@gatech.edu}}%
        \thanks{$^{3}$Thomas Gurriet is a Graduate Research Assistant in Decision and Control Laboratory at the Georgia Institute of Technology, North Ave NW, Atlanta, GA 30332, USA
        {\tt tgurriet3@gatech.edu}}%  
        \thanks{$^{4}$Mark L. Mote is a Graduate Research Assistant of the Decision and Control Laboratory at the Georgia Institute of Technology, North Ave NW, Atlanta, GA 30332, USA
        {\tt mmote3@gatech.edu}}%        
\thanks{$^{5}$Eric Feron is the Dutton/Ducoffe Professor at the School of Aerospace Engineeringat the Georgia Institute of Technology, North Ave NW, Atlanta, GA 30332, USA, {\tt feron@gatech.edu}}
}

\begin{document}

\maketitle
\thispagestyle{empty}
\pagestyle{empty}

%%%%%%%%%%%%%%%%%%%%%%%%%%%%%%%%%%%%%%%%%%%%%%%%%%%%%%%%%%%%%%%%%%%%%%%%%%%%%%%%
\begin{abstract}
This paper describes the process and challenges behind the design and development of a microgravity enabling aerial robot. The vehicle, designed to provide at minimum 4 seconds of microgravity at an accuracy of \textbf{$10^{-3}$} g's, is designed with suggestions and constraints from both academia and industry as well a regulatory agency. The feasibility of the flight mission is validated using a simulation environment, where models obtained from system identification of existing hardware are implemented to increase the fidelity of the simulation. The current development of a physical test bed is described. The vehicle employs both control and autonomy logic, which is developed in the Simulink environment and executed in a Pixhawk flight control board.   
\end{abstract}

%%%%%%%%%%%%%%%%%%%%%%%%%%%%%%%%%%%%%%%%%%%%%%%%%%%%%%%%%%%%%%%%%%%%%%%%%%%%%%%%
\section{Motivation}
\begin{figure}[h]
\begin{center}
\includegraphics[width=8.5cm]{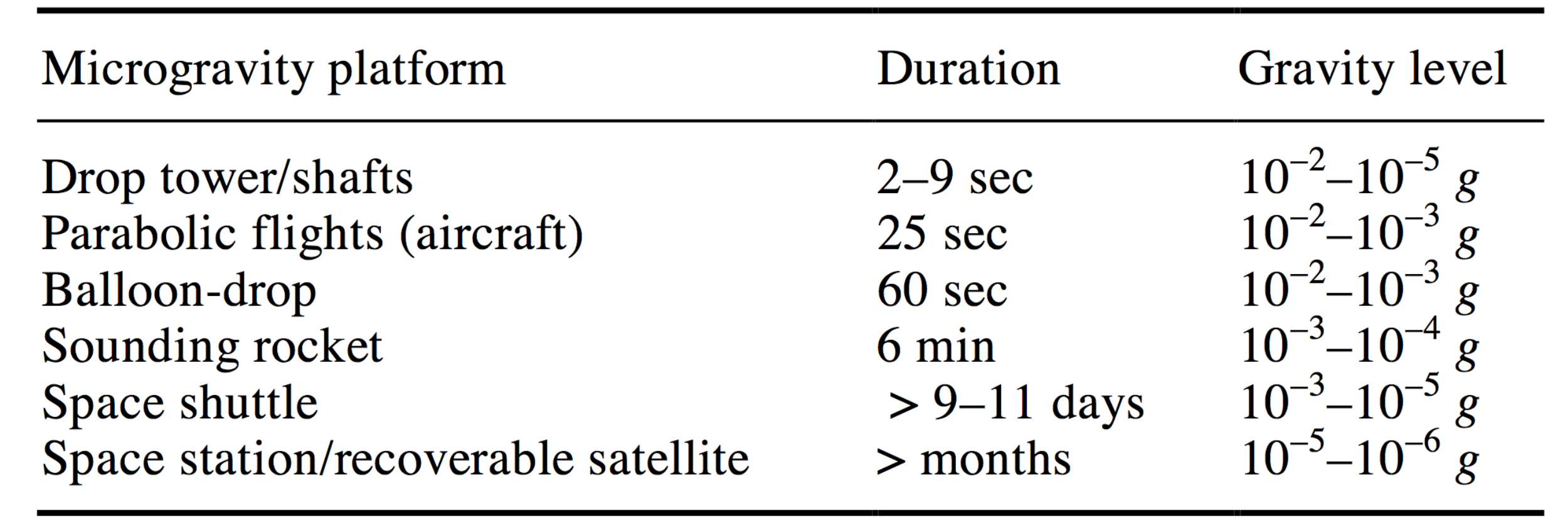}
\caption{Comparison of the available microgravity research platforms \cite{Prasad}}
\label{ZGTable}
\end{center}
\end{figure}

The properties of microgravity conditions make it a valuable environment to conduct research otherwise unfeasible. Fields as diverse as materials science, fluid physics, combustion, biology and biotechnology all have current research questions proposed relating to microgravity conditions~\cite{flies_zg,crystals,combust}. The microgravity research being conducted around the world is already proving its usefulness in challenging and validating contemporary scientific theories as a result of unexpected or unexplained discoveries. Since 1981, on average approximately \$100 million are being spent annually towards microgravity research by space agencies in America, Europe, Germany and Japan \cite{Proceedings}. Most of this research could be conducted in the long term microgravity conditions the International Space Station provides in orbit. However, for most researchers, access to space-based microgravity is prohibitively expensive, to the point of being entirely out of reach. Furthermore, many of these experiments only require brief periods in microgravity, on the scale of a few seconds. For these reasons, a large amount of work has been done enabling terrestrial microgravity options. Fig. \ref{ZGTable} provides an overview of the currently available microgravity platforms and their corresponding details. 
%However, most of these methods have a drawback that can
%hamper access by researchers in one form or another.

\begin{figure}[h]
\begin{center}
\includegraphics[width=8cm]{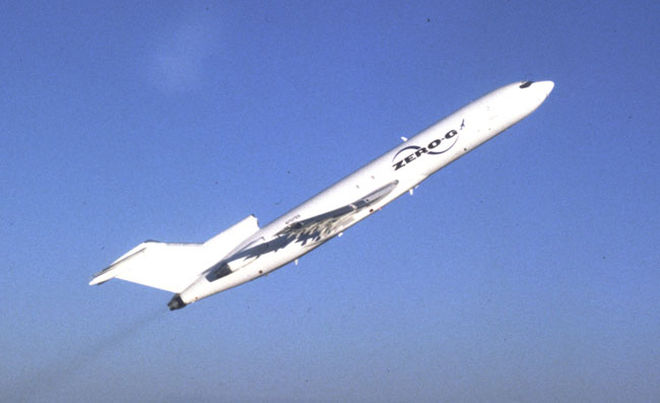}
\caption{Vomit Comet beginning a microgravity parabola \cite{vc_cite}}
\label{vc}
\end{center}
\end{figure}

Two of the most commonly used terrestrial based options to generate microgravity are parabolic manned aircraft and drop towers. The “Vomit Comet”, an aircraft capable of parabolic flight resulting in weightlessness conditions within the fuselage is perhaps the most common example of this technology \cite{vomit}. A Boeing 727 was NASA's "Vomit Comet" for years; it is now operated by the Zero Gravity Corporation (cf. Fig. \ref{vc}).

As illustrated by Fig. \ref{ZGTable}, the Vomit Comet can provide microgravity for up to 25 seconds with a gravity level threshold of $10^{-2}$-$10^{-3}$ $g$'s. This method has the drawback of high price, which has been recorded to be as high as \$3,000 USD/kg \cite{Prasad}. Another major draw back is the lack of repetition, as the aircraft is often booked months in advance and can be very selective \cite{ESA}. Drop towers can potentially solve the issues arising with the parabolic flight. They are cheaper per use than the parabolic aircraft and have a very fast turn around time. However, the main drawback with drops towers is the large upfront cost of necessary infrastructure. Furthermore, the expected microgravity time is on the lower end of the spectrum, usually limited by the height of the tower. \cite{Haber}. Hence, in order to fulfill the current gap faced by researchers from a variety of fields, a quad-rotor-based platform requiring a significantly lower investment, with increased availability and reliability, is now discussed.

The recent explosion of the Unmanned Aerial Vehicle (UAV) market has enabled the development of high end hobby grade components that are capable of creating 0-$g$ conditions. Therefore, we propose the development of such a vehicle with the potential of extending the field of microgravity experiments. This microgravity UAV option is not intended to replace any existing platforms, but rather provide researchers with a tool in which experiments can be refined enough to warrant the investment to the more costly and selective platforms.   
  
\section{Design Methodology}

The adopted design methodology suggests by starting with a clear description of the problem under consideration, in which the project requirements and constraints are stated and elaborated. The motivation section in this paper is clear in describing the need behind available microgravity platforms. Hence, the necessary requirements and constraints are now described in detail.

\subsection{Requirements and constraints}
A product discovery process, led by the research team and composed of interviews performed at multiple schools within Georgia Institute of Technology, revealed the requirements and constraints involved in the the development of a microgravity platform. These are now itemized and described below. 

\begin{itemize}
  \item Affordability: Vehicle must cost less than \$25,000 USD,
  \item Reliability: Vehicle must be produce repeatable results, 
  \item Time Exposure: Vehicle must be able to provide at minimum $5$ seconds of microgravity,
  \item Regulatory Constraints: Vehicle must not violate regulations realtive to the operation of civilian drones.
\end{itemize}

The decision to use an autonomous UAV was made since it fits perfectly within the requirements and constraints discovered during the product discovery process. Furthermore, the choice of a quad rotor was made since it solves several drawbacks from current technologies available. First, an autonomous quadrotor does not require any supporting infrastructure, so it can be deployed and operated almost everywhere. Since it is capable of vertical flight, the real state necessary for testing is minimal. Furthermore, federal regulations posed by the US Federal Aviation Administration permits its use as long as it remains within eyesight, under 400ft (approximately 120 meters) and with a mass lower than 55 pounds (approximately 25 Kilograms). In Germany, the maximum allowable altitude is 500 feet and maximum mass is 25 Kilograms.
From a financial standpoint, quadrotors are mechanically simple, which makes them available and affordable to any academic research institution.

\subsection{Preliminary design}

\begin{figure}[h]
\begin{center}
\includegraphics[width=8cm]{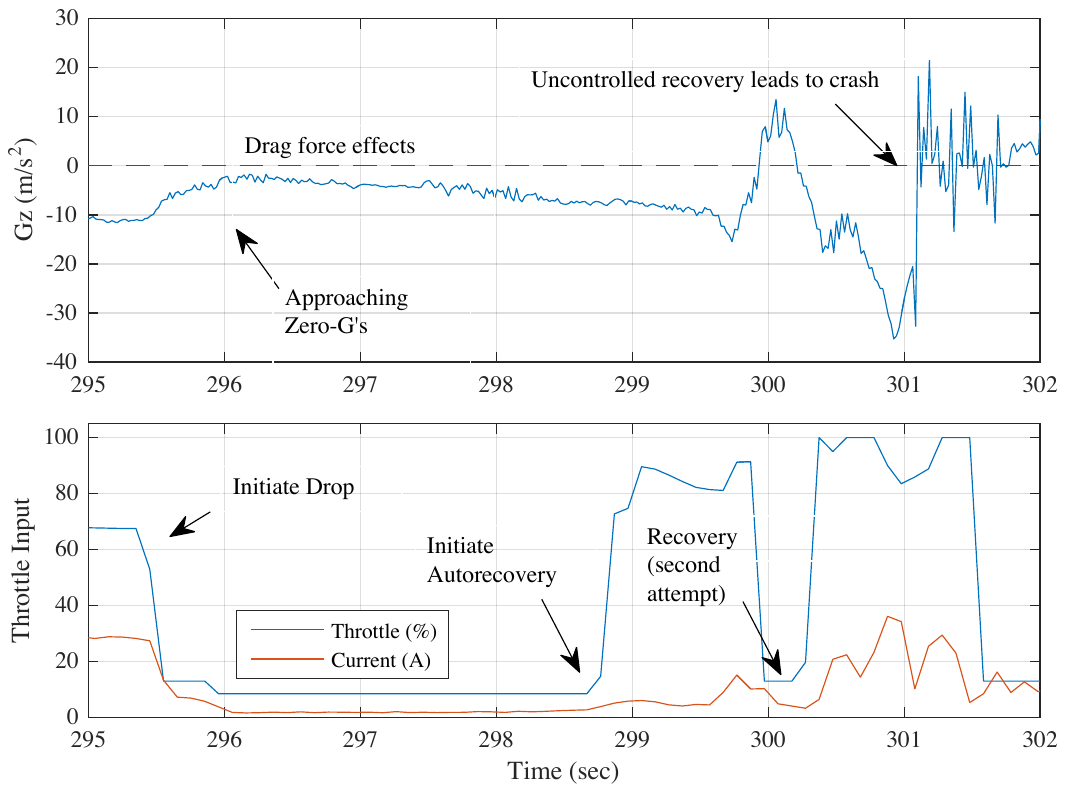}
\caption{Fixed pitch trajectory}
\label{fp}
\end{center}
\end{figure}

Several modifications on the conventional quadrotor architecture are necessary to obtain a vehicle capable of producing microgravity. First, tracking a microgravity trajectory requires the use of both positive and negative thrust forces in order to compensate aerodynamic drag in ascent and descent portions of the parabolic flight; something that a fixed pitch quadrotor cannot do. Furthermore, since the thrust profile on a fixed pitch quadrotor is a quadratic function of the propeller rotational velocity, the control authority tends to zero as the thrust required goes to zero and hence, vehicle instability occurs. Fig. \ref{fp} shows experimental data for a fixed pitch quadrotor attempting to track a microgravity profile. 

Note that both slow responsiveness of fixed pitch electric motors, as well as aerodynamic drag forces prevent the vehicle from reaching the zero gravity line. Furthermore, since the thrust forces are nearly zero, the vehicle is incapable of retaining proper attitude during the maneuver and the results are catastrophic.

To overcome the issues observed during the fixed pitch quadrotor experiments, the research team developed a variable pitch quadrotor. This modification, which also results in using constant rotor speed, would enable a more responsive system that is capable of (i) fighting drag independent of its direction and (ii) maintaining attitude control authority independent of the thrust required during a microgravity tracking flight.

\subsection{Requirement based Sizing}

\begin{figure}[H]
\begin{center}
\includegraphics[width=8cm]{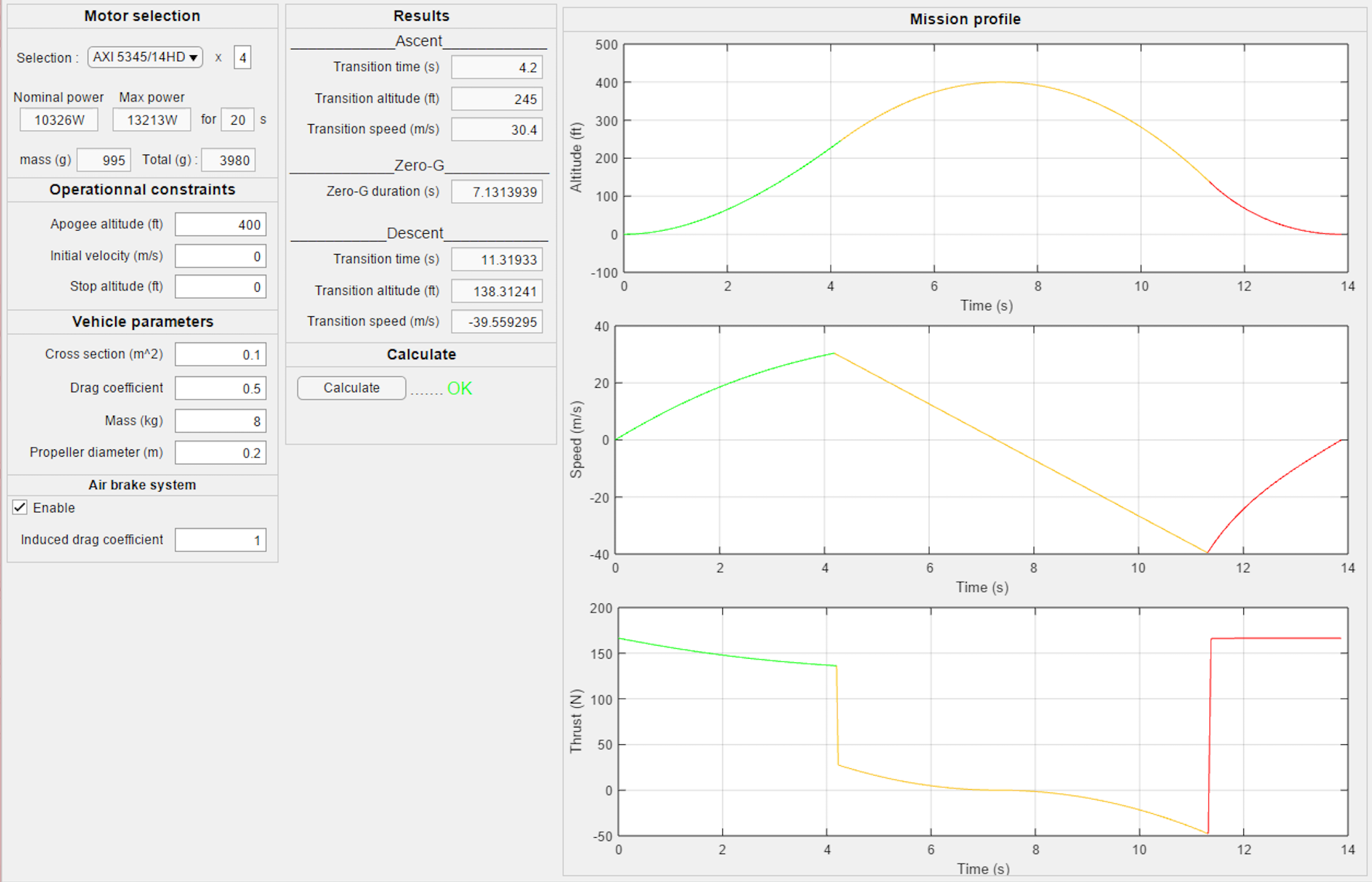}
\caption{Custom Sizing Tool}
\label{app}
\end{center}
\end{figure}

Since one of the primary goals of this project is maximizing the time spent in microgravity, while staying within UAV operational limits, the authors developed an interactive sizing tool. The tool employs a physics-based approach along with the use of manufacturer data to enable the study of the interplay between vehicle size and mission profile. The tool employs Matlab appDesigner \cite{matlab_appD} and allows to the interactive definition of all the constraints and parameters that compose the problem. Hence, using user inputs, the tool computes a trajectory that optimizes the time spent in microgravity. Profiles of altitude, velocity and thrust enabling the trajectory are then suggested, which allow for an educated decision when sizing the vehicle, as illustrated in Fig. \ref{app}.

For the preliminary generation of this optimal trajectory, the vehicle is modeled as a one dimensional system traveling along a vertical axis. Hence, the equations of motion are reduced to a second-order non-homogeneous differential equation given by Eq.~(\ref{appeq1})

\begin{equation}
\ddot{h}=\frac{F}{m},
\label{appeq1}
\end{equation}
where the forces modeled include aerodynamic, gravitational and propulsive components. Gravitational $F_d$ and aerodynamic $F_a$ components of the force vector are given by Eqs.~(\ref{2.1}) and~(\ref{2.2}).

\begin{equation}
F_{g}=-mg, 
\label{2.1}
\end{equation}
\begin{equation}
F_{d}=\frac{1}{2}\rho(h)S C_{d} \dot{h}^{2}, 
\label{2.2}
\end{equation}
where $m$ is the vehicle's mass, $g$ is gravity, $S$ is the planform area, $\rho(h)$ is the air density and $C_d$ is the drag coefficient. Since there is not an analytical expression for thrust forces, and propeller characteristics are hard to find for commercially available hobby products, elementary momentum theory of propellers is used to get the ideal thrust as a function of speed; it is given by Eq.~(\ref{2.3})
\begin{equation}
T(\dot{h})=\frac{\eta P_{\rm engine}}{\dot{h}}, 
\label{2.3}
\end{equation}
where the efficiency $\eta$ is the solution 
of Eq.~(\ref{2.3.1}) 
\begin{equation}
\dot{h}=\eta \left(\frac{2 P_{\rm engine}}{\pi \rho(h) D^{2} (1-\eta)}\right)^{\frac{1}{3}}. \tag{4.1}
\label{2.3.1}
\end{equation}
Note that for the special case of static thrust,
\begin{equation}
T\left(0\right)=\frac{\pi}{2}D^{2}\rho(h) P_{\rm engine}^{2}, \tag{4.2}
\end{equation}
Where it is assumed that the propeller control system optimizes the blade
pitch for maximum efficiency, with the power $P_{\rm engine}$ generated by the motor 
as the control input. Here $D$ is the propeller diameter. Ultimately, this ideal thrust is reduced by a factor
of 0.7 to account for all the dissipative forces in play and be more accurate with respect to real propellers efficiency.

\begin{figure}[h]
\begin{center}
\includegraphics[width=8cm]{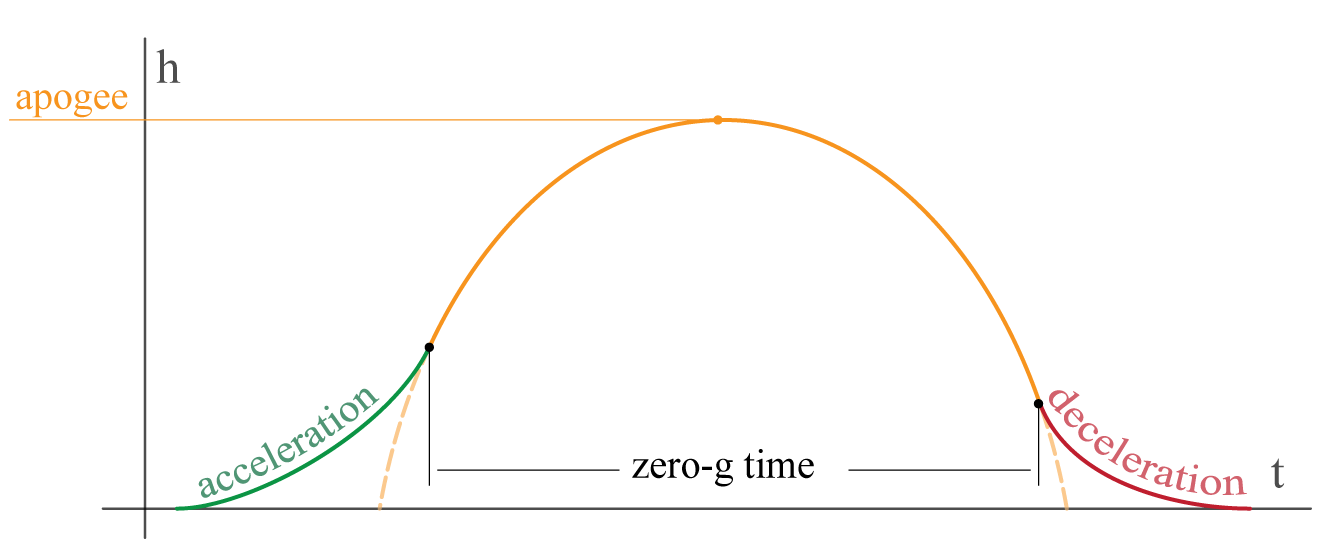}
\caption{Trajectory generation}
\label{traj}
\end{center}
\end{figure}

Once in place, the challenge becomes finding a sequence of control inputs that maximizes the time in microgravity subjected to the previously defined dynamics and constraints. Hence, we use the maximum altitude allowed (400 ft), the minimum altitude at which to stop during the descent, and maximum power available based on data-sheets from \cite{axi_motors}. The use of a launching mechanism can be taken into account by adjusting the initial velocity at takeoff, while the use of air-brakes can be taken into account by adjusting values for the coefficient of drag $C_d$ during the braking phase of the mission. A descriptive illustration of the mission is provided by Fig. \ref{traj}.

An optimal solution to this problem can be obtained by employing any non-linear optimizer; However, the solution to this problem can be found from elementary considerations. For example, it was observed that as long as the altitude requirement is not violated, the time spent in microgravity, during the acceleration section of the mission (shown by a green line in Fig. \ref{traj}), is a monotonically increasing function of both thrust available and vehicle acceleration. Similarly, thrust available and the employment of air-brakes maximizes microgravity during the deceleration portion of the mission as a monotonically increasing function of the apogee altitude of the 0-$g$ parabola. Hence, the optimal solution an intuitive "bang-coast-bang" policy \cite{BCB}where maximum power is applied until the UAV merges with the parabola whose apogee is the maximum allowed altitude, and then begins breaking with maximum intensity as late as possible so as to stop at the desired "park" altitude, which is bounded below by the ground. The previous considerations imply that in terms of calculating the actual trajectory, one must only find the time at which these two transitions must occur based on vehicle capabilities. Hence, a simple numerical integration of the equations of motions is used in conjunction with an iterative solver based on Newton's method to find the 2 transition events marking the beginning and the end of the 0-$g$ flight segment. 

\subsection{Physical modeling}
\begin{figure}[h]
\begin{center}
  \includegraphics[width=8.5cm]{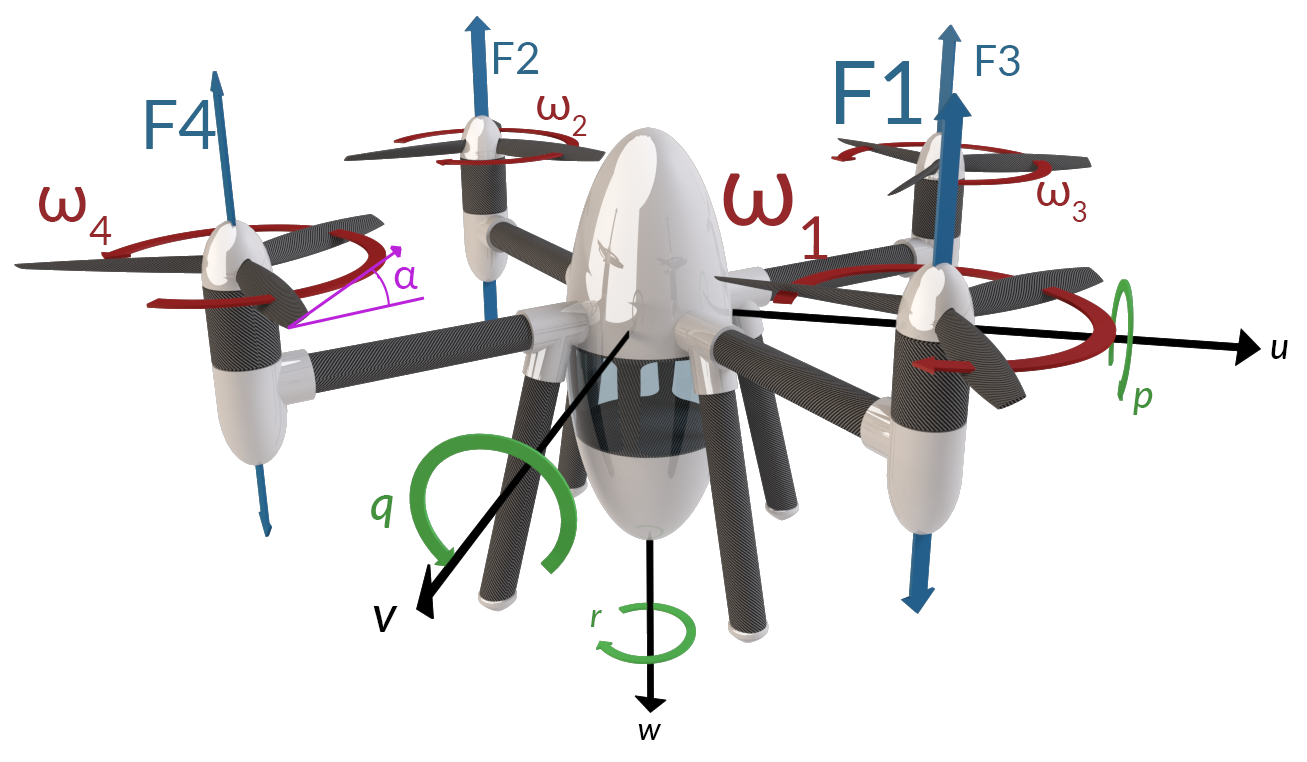}
  \caption{Diagram of variable pitch quadrotor concept illustrating commonly used notation}
  \label{render}
  \end{center}
\end{figure}

Once the problem is properly defined and requirements are in place, a full 6 DOF model is derived. We begin by introducing the linear kinematic equations of motion for a quadrotor, which are given by the Newton-Euler equations~(\ref{linkin}). 
\begin{equation}
\begin{split}\label{linkin}
\dot{u}&=vr-wq-g\rm{sin}\theta + F_x/m \\
\dot{v}&=wp-ur+g\rm{sin}\phi \rm{cos}\theta + F_y/m\\
\dot{w}&=uq-vp+g\rm{cos}\phi \rm{cos}\theta + F_z/m
\end{split}
\end{equation}
Note that the previous equations employ notation from the linear velocity vector $\vec{U}=[u,v,w]^T$. Similarly, the angular kinematic equations of motion are given by the Newton-Euler equations~(\ref{angkin}); the cross products of inertia are neglected.
\begin{equation}
\begin{split}\label{angkin}
\dot{p}&=qr(I_{yy}-I_{zz})/I_{xx} + M_x/I_{xx}\\
\dot{q}&=pr(I_{zz}-I_{xx})/I_{yy} + M_y/I_{yy}\\
\dot{r}&=pq(I_{xx}-I_{yy})/I_{zz} + M_z/I_{zz}
\end{split}
\end{equation}
Note that the previous notation is composed of components from the angular velocity vector $\vec{\Omega}=[p,q,r]^T$ and the inertia tensor.
The set of forces and moments acting on the system, denoted by $\vec{F}=[F_x,F_y,F_z]$ and $\vec{M}=[M_x,M_y,M_z]$ respectively, are organized as
\begin{equation}
\begin{split}\label{x}
\vec{F} &= F_{{gravity}} + F_{{propulsion}} + F_{{aerodynamic}}\\
\vec{M} &= F_{{drag-torque}} + F_{{propulsion}} 
\end{split}
\end{equation}
Fig.~\ref{render} illustrates the commonly used variables necessary to describe a variable pitch quadrotor.

The rotational kinematic equations were mechanized using quaternions \cite{quat}. The inertial velocities are derived from the body-axis velocities by a coordinate transformation (flat-Earth equations are used) and integrated to obtain inertial position. A fourth-order Bogacki-Shampine integration method \cite{matlab_solve} is used, with an integration step of 0.004 seconds.

\subsection{Simulation Environment}
\begin{figure*}[!b]
  \includegraphics[width=\textwidth]{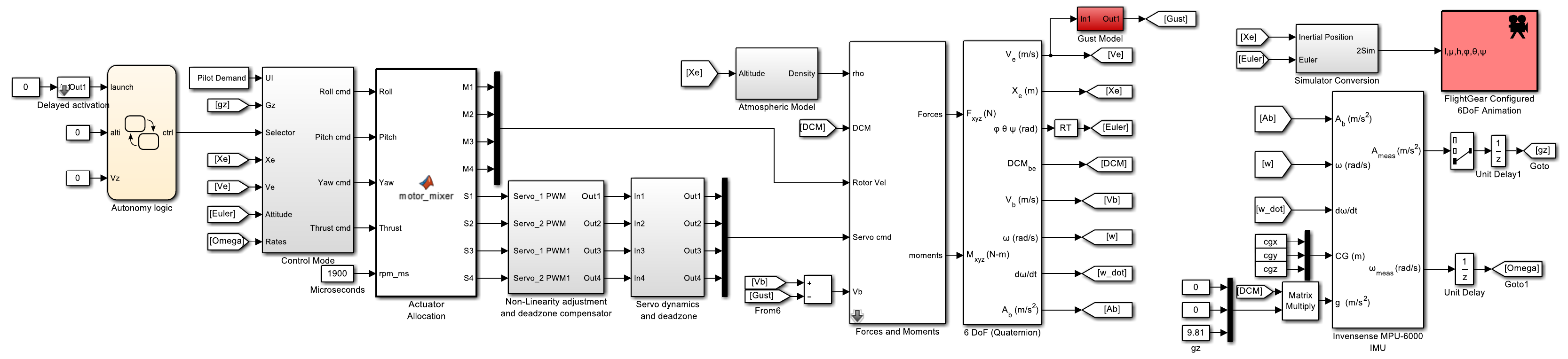}
  \caption{Model Based Design in the Simulink Environment}
  \label{mbd}
 \end{figure*}

The development of a model inside of a simulation environment enables the evaluation of modeling assumptions and control algorithms. Fig. \ref{mbd} is a high level illustration of the model developed inside the Simulink environment. 

In order to visualize the model properly, a direct link is established between Simulink and FlightGear. A link to a video of the Simulink-FlightGear simulation is provided on the caption of Fig. \ref{simgz}

\subsection{Attitude Control}
\begin{figure}[H]
\begin{center}
\includegraphics[width=8cm]{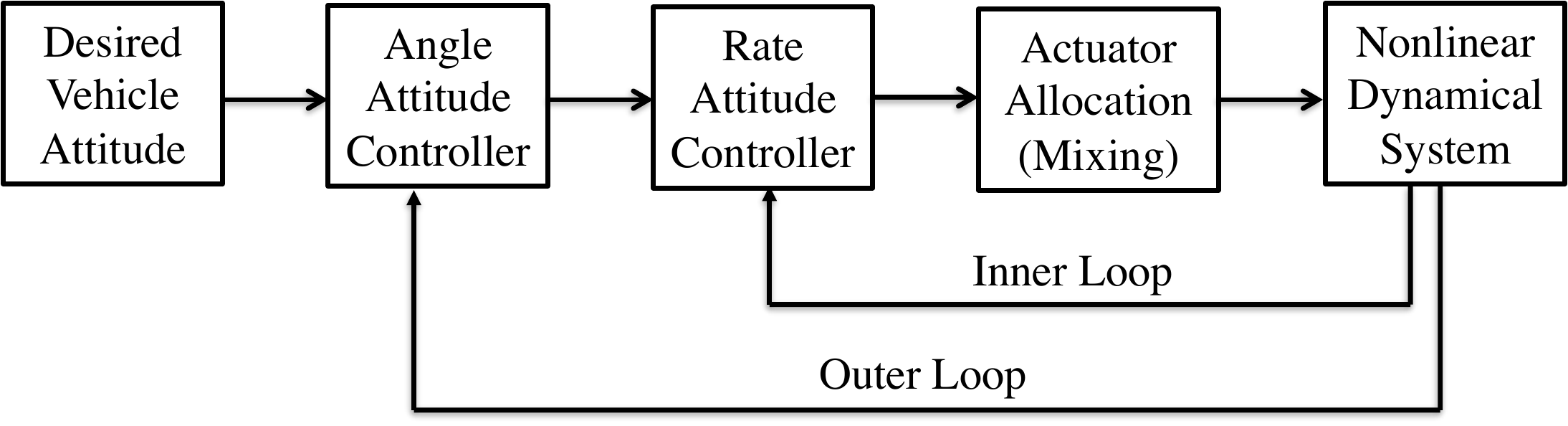}
\caption{Nested PID Control Algorithm}
\label{nest}
\end{center}
\end{figure}

Once the characteristics of the model are accounted for, the next step aims at the derivation of a control algorithm that will be executed by the embedded platform. The attitude control law employed in the vehicle is commonly known as a cascade feedback controller. After a thorough review of control algorithms for autonomous quadrotors \cite{zulu}, the PID cascade control strategy was chosen due to its simplicity, precision, tracking ability, fast convergence and robustness. Fig.~\ref{nest} provides a high level overview of this controller.

\begin{figure}[H]
\begin{center}
\includegraphics[width=8cm]{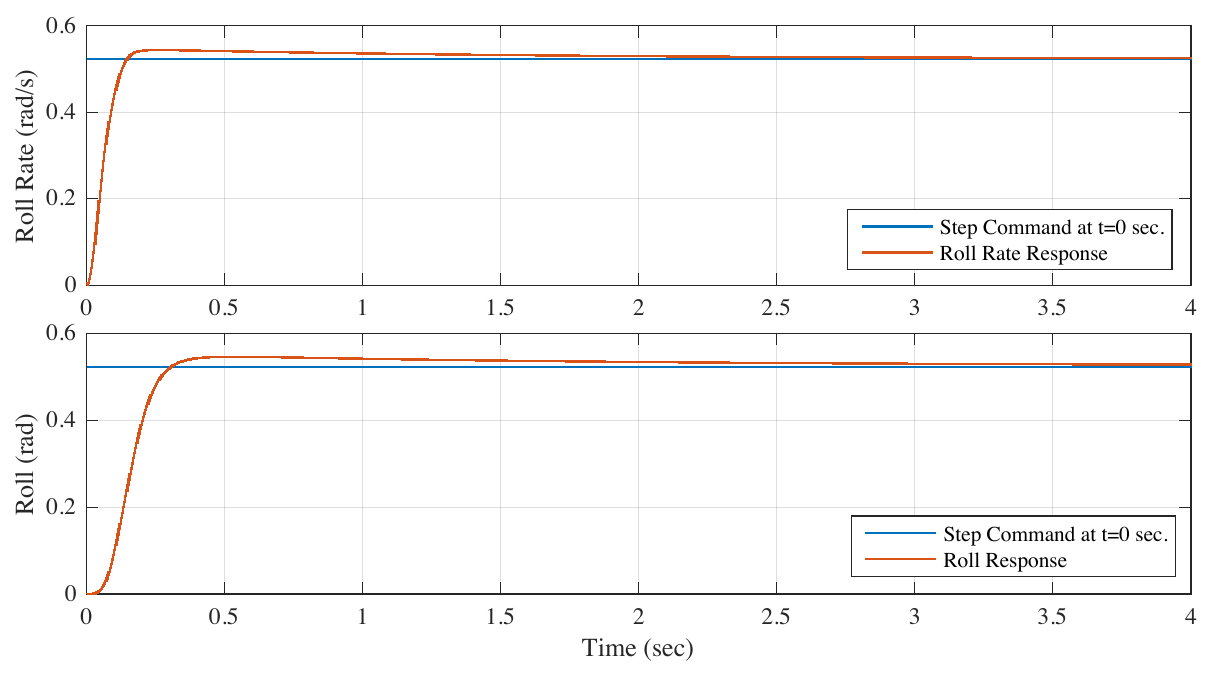}
\caption{Controlled response of roll and roll rate to step input}
\label{response}
\end{center}
\end{figure}

In control theory, the PID controller in a parallel structure is represented in the continuous time domain as 
\begin{equation}
u(t) = K_pe(t)+K_i\int_{t_i}^{t_f}e(t)\textrm{d}\tau+K_d\frac{\textrm{d}e(t)}{\textrm{d}t},
\end{equation}
where each subscript represents the proportional, integral and derivative component of the three gains denoted by $K_p$, $K_i$ and $K_d$ respectively.
Gain tuning was performed after a linearized model was obtained. First, vehicle rates were tuned such that proper tracking and risetime was achieved. Once the rate responses met a satisfactory response, a the angle attitude was tuned similarly with the rate loop activated. Responses of tuned roll dynamics are illustrated in Fig. \ref{response}

\subsection{Actuator Allocation}
The actuator allocation block illustrated in Fig. \ref{nest} block-diagram computes the required blade pitch of each variable pitch system in order to generate the appropriate forces and moments required to stabilize and navigate the vehicle. The linear relationship between the forces and moments to the blade pitch of each propeller is given by Eq.~(\ref{mat}).
\begin{equation}
\begin{bmatrix}
T\\
L\\
M\\
N
\end{bmatrix}
=
\begin{bmatrix}
   K_{T} & K_{T} & K_{T} & K_{T} \\
   -K_{T}L_Y & -K_{T}L_Y & K_{T}L_Y & K_{T}L_Y \\
   K_{T}L_X & -K_{T}L_X & -K_{T}L_X & K_{T}L_X\\
    K_{D} & -K_{D} & K_{D} & -K_{D}
\end{bmatrix}
\begin{bmatrix}
\alpha_1\\
\alpha_2\\
\alpha_3\\
\alpha_4
\end{bmatrix},
\label{mat}
\end{equation}
where T indicates the thrust forces, L, M, N the moments about the x, y, z, axes respectively for each of the individual blade deflections $\alpha_i$. Furthermore, $K_T$, $K_D$. $L_X$ and $L_Y$ are the coefficients that make this linear mapping possible.
%Note that in order to determine the blade pitch required from each propeller corresponding to a given thrust force and moment, the inverse of this relationship is necessary, but not shown in this work.

Since the acceleration phase of the mission shown in Fig. \ref{traj} requires maximum thrust available in order to maximize microgravity time, actuator inputs are set to their maximum value. However, this interferes with the vehicles attitude control law and would leave the vehicle incapable of correcting itself since actuators are maxed out at this point. Hence, the derivation and implementation of an enhanced actuator allocation strategy was performed. This modification allows for maximum total thrust while ensuring that attitude controller is capable of operating by sizing the input signals when necessary such that saturation never occurs. This approach is based on the observation that for a variable pitch quadrotor, pitch and the roll are affected by the difference in thrust on the opposites pairs of motors. The derived algorithm is provided by Alg. 1.

\begin{algorithm}
\caption{Mixing algorithm}\label{euclid}
\begin{algorithmic}[1]
\Procedure{Mixing}{$R_{cmd}, P_{cmd}, Y_{cmd}, T_{cmd}$}
\State $\textit{O} \gets \textit{$(R_{cmd}, P_{cmd}, Y_{cmd}, T_{cmd})$}$\\

\State $\textit{K} \gets \left[\begin{array}{cccc}
-1 & 1 & 1 & 1\\
1 & -1 & 1 & 1\\
1 & 1 & -1 & 1\\
-1 & -1 & -1 & 1
\end{array}\right]$\\

\State $U \gets K*O$

\If {$U$ \text{saturates}}	
\If {$yaw_{cmd} \neq 0$} \Comment{Improvement possible}
\State Find the biggest \textit{$\alpha \in [0,1]$} such that
\State $ U=(R_{cmd}, P_{cmd}, \alpha*Y_{cmd}, T_{cmd})$ 
\State doesn't saturate

\If {such an $\alpha$  exists}
\State \Return $ K*(R_{cmd}, P_{cmd}, \alpha*Y_{cmd}, T_{cmd})$
\EndIf

\ElsIf {$T_{cmd} \neq 0$}
\State Find the biggest \textit{$\alpha \in [0,1]$} such that
\State $ U=(R_{cmd}, P_{cmd}, 0,  T_{cmd})$ doesn't saturate
\If {such an  $\alpha$ exists}
\State \Return $K*(R_{cmd}, P_{cmd}, 0, \alpha*T_{cmd})$
\EndIf

\Else
\State Find the biggest \textit{$\alpha \in [0,1]$} such that
\State $U=(R_{cmd}, P_{cmd}, 0, 0)$ doesn't saturate
\State \Return $K*(R_{cmd}, P_{cmd}, 0, 0)$
\EndIf
\Else
\State \Return \textit{K*O}

\EndIf
\EndProcedure
\end{algorithmic}
\end{algorithm}

The proposed actuator allocation algorithm ensures that the commanded roll $R_{cmd}$ and pitch $P_{cmd}$ are the last to hit the saturation. Hence, roll and pitch have the highest priority, followed by thrust, and lastly yaw.  

\subsection{System Identification of actuators}
\begin{figure}[h]
\begin{center}
\includegraphics[width=8cm]{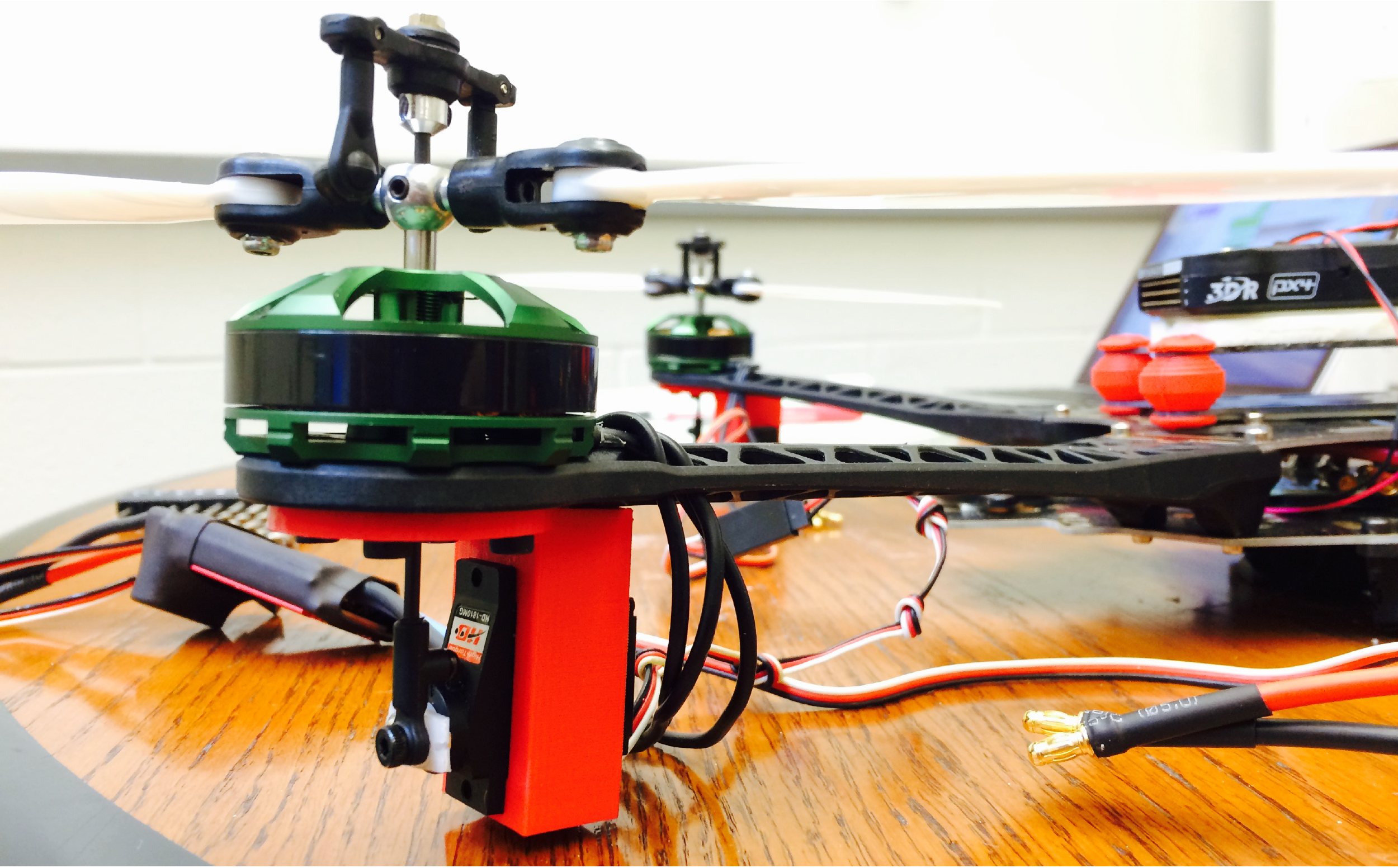}
\caption{Variable pitch mechanism with custom 3D-printed servo mounts}
\label{vp}
\end{center}
\end{figure}

In order to increase the fidelity of the simulation, system identification was performed on the variable pitch system illustrated by Fig. \ref{vp}.

\begin{figure}[h]
\begin{center}
\includegraphics[width=8cm]{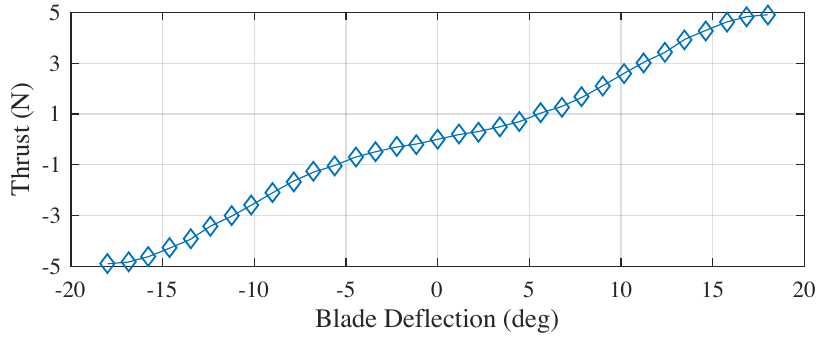}
\caption{Static thrust curve for blade deflection envelope }
\label{thrust}
\end{center}
\end{figure}

This task enabled understanding of the servo dynamics, as well as the relationship between blade pitch and thrust for a given motor speed. The command ranges for a given blade pitch are symmetric around the neutral point. The maximum commanded deflections were set in radians as $\delta_r = \pm 0.09$. Actual maximum pitch angles of the rotor blades were measured statically. Furthermore, linear transfer functions are used to model the servo dynamics. The HD-1810MG servos used in the variable pitch mechanism were subjected to small-amplitude frequency sweeps while attached to a load cell. Since the system was fully operational, it was assumed to be representative of the actual loads experienced by the servos during the flight. The following transfer function was obtained using the experimental data.
\begin{equation}
H_{servo}(s)=\frac{1}{\tau s+1}, 
\end{equation}
where $\tau = 0.075$ seconds. A dead-band of 5 micro-seconds was also applied to account for servo backlash as suggested by the manufacturer data-sheet \cite{servo}. A static thrust experiment sweeping through the envelope of interested revealed a non-linear static thrust curve, see Fig. \ref{thrust}. Thus, in order to enforce linearity on the available thrust, the thrust curve was linearized through static inversion.

\subsection{Automation}
\begin{figure}[h]
 \begin{center}
\includegraphics[width=8cm]{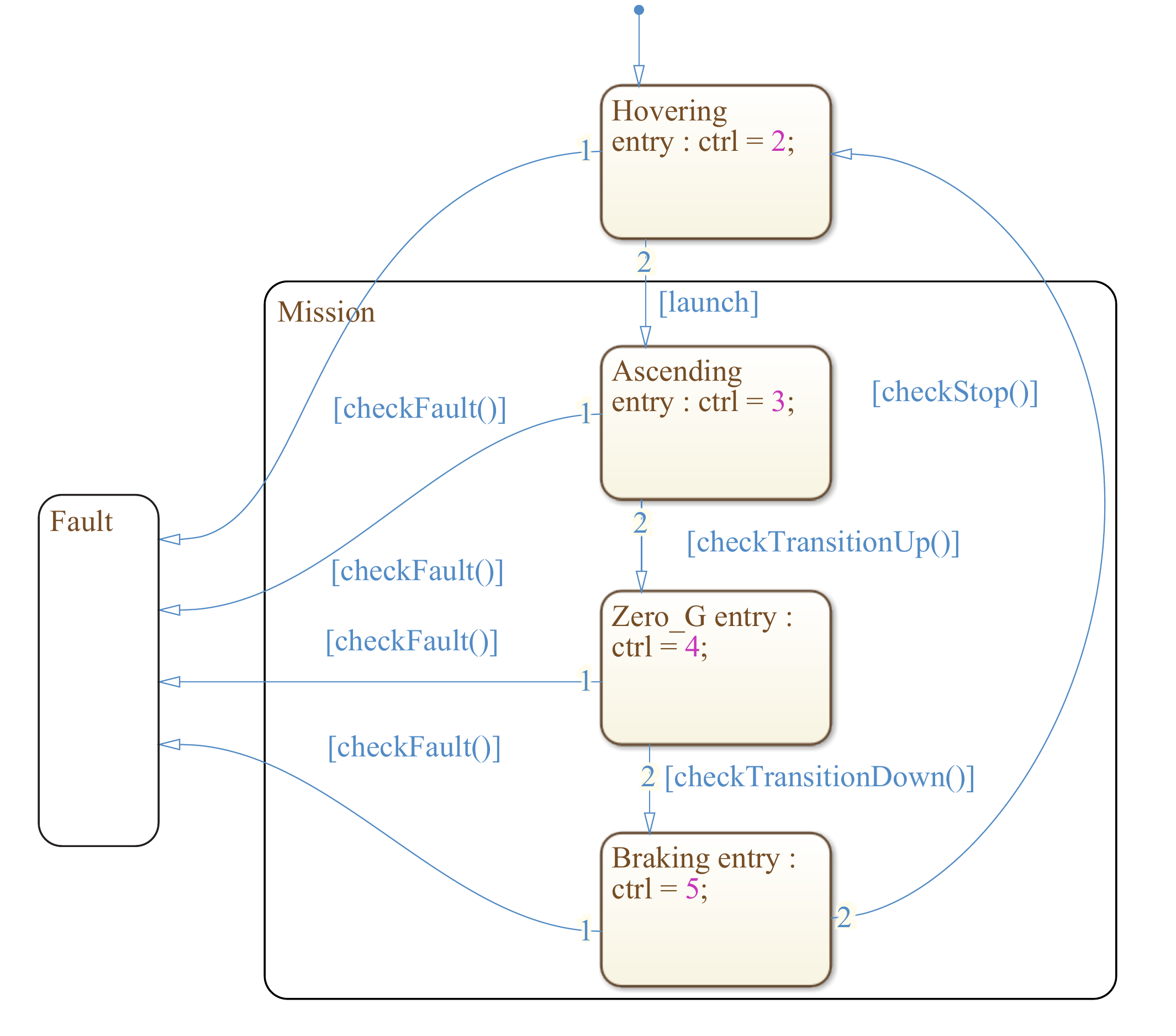}
\caption{Autonomous logic represented as StateFlow diagram \cite{state_flow_doc} }
\label{stateflow}
  \end{center}
\end{figure}

%This section overviews the control framework needed to generate the set of necessary positions $\vec{x}_E^T=[x_{E,d}, y_{E,d}, z_{E,d}]$ and vehicle attitudes  $\vec{\Phi}^T_E=[\phi_{E,d}, \theta_{E,d}, \psi_{E,d}]$ as specified in the Earth fixed (north-east-down) coordinate frame $\mathcal{F}_E$. 
An autonomous flight mode was designed for precise execution of the microgravity maneuver, while a manual flight mode enables the pilot to gain control of the vehicle. During the manual flight mode, thrust magnitude and attitude angle commands from the pilot are sent via Radio Controller (RC). %The autonomous mode employs a trajectory generator along with PID controllers on the outermost loop to generate the position command $\vec{x}_E^T$. 
The three phase time optimal trajectory plan for microgravity tracking is initiated through a manual switch. The autonomous controller switches between three sub-modes in its autonomous configuration, as outlined in Fig. \ref{stateflow}. These switches during an overall aggressive flight sequence are reminiscent of prior work by the authors~\cite{PiF:01}. The ascent phase is initiated with a five second time delay and warning sound from the vehicle, after which maximum thrust is commanded until the microgravity parabola is entered and the vertical acceleration controller for the microgravity management takes over. During this phase, a controller compensates the error of the filtered vertical acceleration measurement signal with thrust output commands. The final stage of the maneuver commands full thrust in order to stop the vehicle descent (and will deploy air-brakes in the next prototype) and stop at the desired altitude. Once this altitude is reached and the vehicle stabilized, the mode switch back to manual flight mode.

%% Need to upgrade the stateflow diagram!!!

\begin{figure}[h]
\begin{center}
\includegraphics[width=8cm]{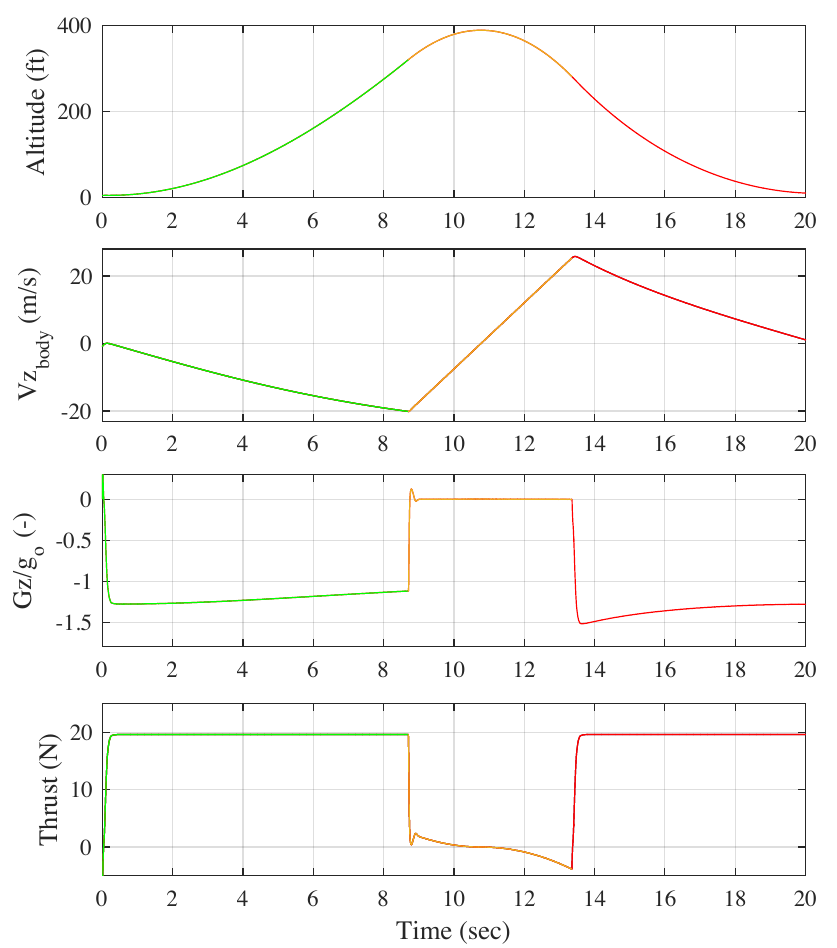}
\caption{Profiles from 6-DOF Simulation \href{https://youtu.be/H4o8Cf8uQ9g}{(Click to see video)}}
\label{simgz}
\end{center}
\end{figure}

Results from the simulation are provided in Fig. \ref{simgz}. In order to prevent the vehicle from drifting due to lateral gusts, a position control algorithm was implemented. Position control in autonomous mode is partitioned into vertical ($z_E$) and lateral ($x_E, y_E$) dimensions.  The position commands are mapped into attitude angle commands and sent to the attitude controller. 

\section{Safety}
\subsection{Online Fault Detection}

\begin{figure}[h]
\begin{center}
\includegraphics[width=8cm]{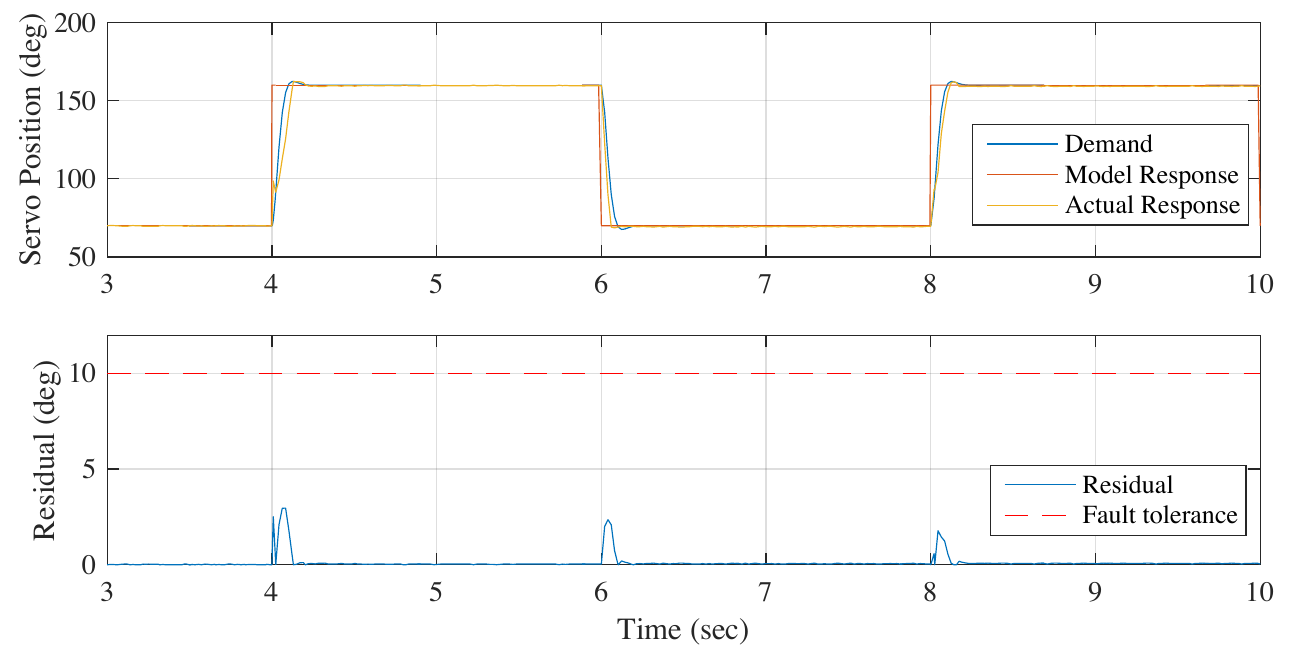}
\caption{Online fault detection employing residual signals}
\label{servoid}
\end{center}
\end{figure}

Control algorithms were developed based on the nominal plant model. Design criteria such as gain and phase margins can ensure acceptable performance when there are slight changes in plant dynamics. However, if plant dynamics change significantly due to a component or sensor failure, the result can be sub-optimal or even catastrophic. To ensure that such failures do not produce such a result, it is important to detect failures as they happen. The execution of an online model enables the comparison of real measurements with predictions from the model and detect failures when the difference exceeds a certain threshold. Models for the servo have been included on board such that if the residual signal r(t), defined as the absolute value of the difference between outputs from the model $y_m(t)$ and actual outputs $y_a(t)$,
\begin{equation}
r(t)=|y_{m}(t)-y_{a}(t)|,
\end{equation}
grows larger than a predefined value, a fault flag is engaged and vehicle is signaled to abort the mission. Fig. \ref{servoid} illustrates this concept using a real servo and a model derived using system identification methods.

\subsection{Geofencing}
\begin{figure}[h]
\begin{center}
\includegraphics[width=8cm]{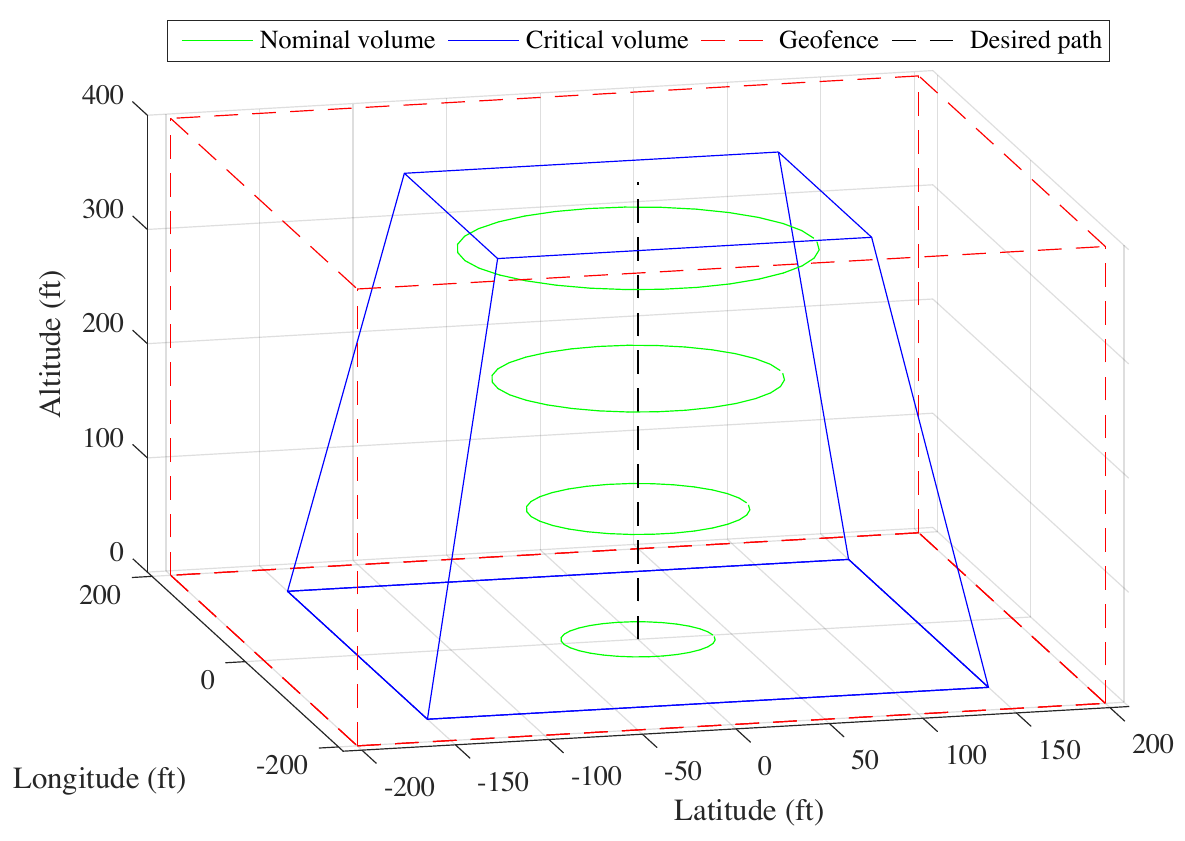}
\caption{Geofence Concept}
\label{geo}
\end{center}
\end{figure}

To reinforce the safety during the mission, a layered geofence structure is defined as depicted of Fig. \ref{geo}. 

First, a "geofence" is defined: The UAV must stay in this safe set of space even in case of failure. Therefore, a "critical volume" is defined as the biggest subset of the geofence such that cutting power to the motors while in that critical volume ensures that the vehicle stays into the geofence. The vehicle will embark an autonomous safety termination system that will cut the power to the motors in case the critical volume is trespassed. Finally, a "nominal volume" is defined as a cone around the nominal path. This volume serves as a geofence that signals an abnormal deviation from the nominal path if the vehicles happens to trespass it. In such a scenario, the current microgravity mission is aborted and a re-centering maneuver is performed to get back to the bottom of the nominal path.

\section{Implementation}
\subsection{Flight test vehicle}
\begin{figure}[h]
\begin{center}
\includegraphics[width=8cm]{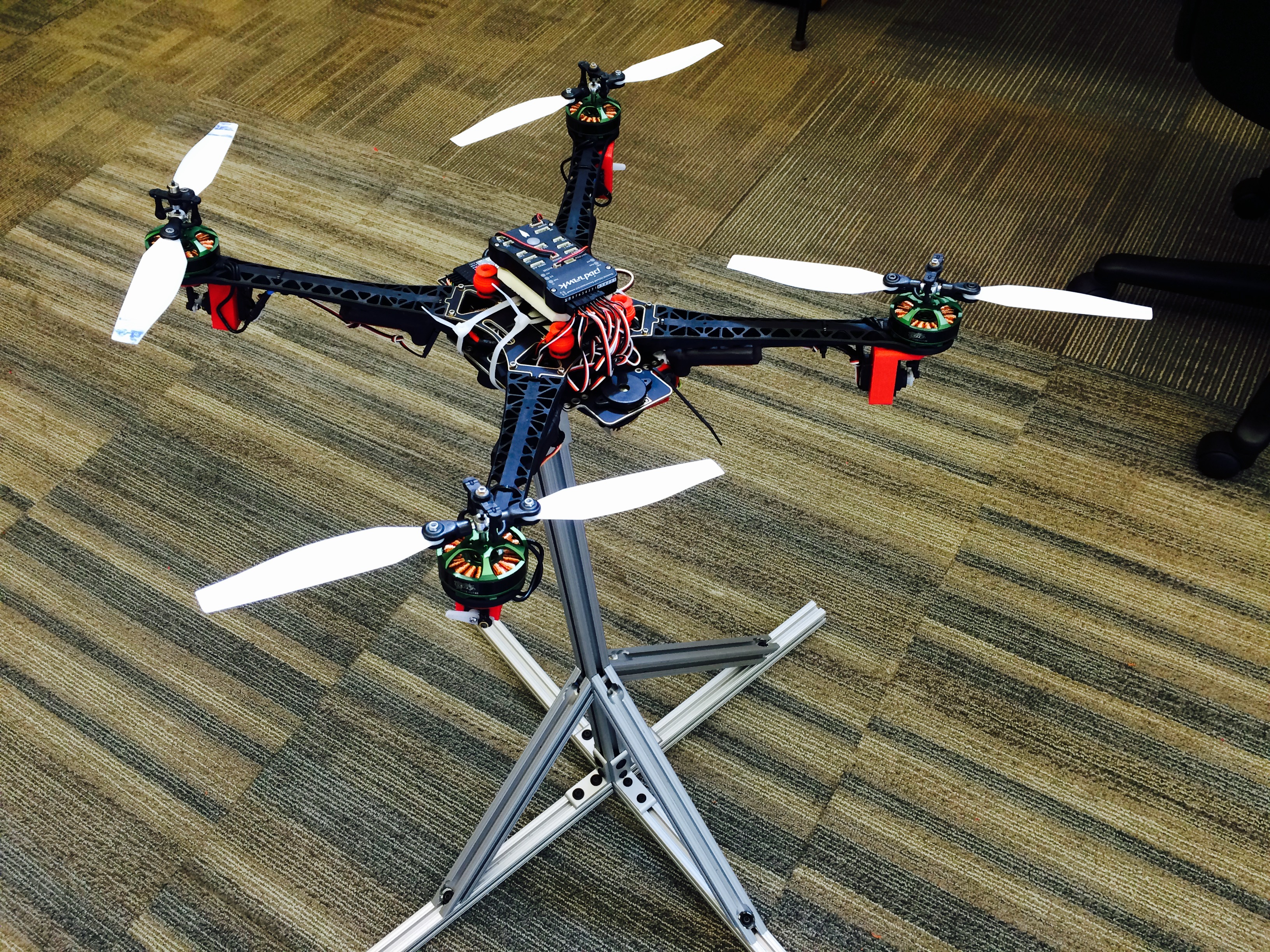}
\caption{Experimental test vehicle on tuning stand}
\label{testbed}
\end{center}
\end{figure}

The model-based development described in the previous sections is used to generate the code employed by the Pixhawk \cite{px4_simulink} flight control board. This hardware was chosen based on its ability to interact with the system and support the automatically generated control algorithms. The Pixhawk flight control board, manufactured by 3D Robotics, features advanced processor and sensor technology from ST Micro-electronics and a NuttX real-time operating system delivering performance, flexibility, and reliability for controlling the vehicle. It contains a dual IMU system where an Invensense MPU 6000 supplements ST Micro LSM303D accelerometer to provide redundancy and improve noise immunity of the power supplies. Based on the detailed specifications, a test vehicle was built and is illustrated through Fig. \ref{testbed}

\subsection{Flight test issues and future work}

The custom-made components and heavily modified motors and variable pitch mechanisms, illustrated in Fig. \ref{vp}, introduced a high level of complexity. Vibrations from unbalanced systems had to be eliminated in order to get operational. Due to these unforeseen challenges, no real microgravity tests have been performed to date. However, the authors plan on disclosing flight test data before the end of September 2016. Although the vehicle has managed to hover, it was observed that the responsiveness of the vehicle could be enhanced by replacing the current set of symmetric blades with some that possess a larger planform area. Due to the lack of availability of symmetric blades and variable pitch systems, the authors are currently manufacturing their own. The final version of this paper will aim at presenting microgravity experimental flight tests results.

\addtolength{\textheight}{-12cm}   % This command serves to balance the column lengths
                                  % on the last page of the document manually. It shortens
                                  % the textheight of the last page by a suitable amount.
                                  % This command does not take effect until the next page
                                  % so it should come on the page before the last. Make
                                  % sure that you do not shorten the textheight too much.

%\section*{APPENDIX}

\section*{CONCLUSION}
The objective of this paper has been to present our design and simulation work towards building a microgravity enabling robot. The reason behind our particular design are explained. The different practical issues arising from this very specific use of a quadrotor UAV are discussed and solutions are proposed. A sizing tool as well as a fully functional 6DoF simulator based on hardware identification data has been developed, allowing us to confirm the viability of the proposed design and control architecture. Our efforts are now focused on implementing this work on a real system and test the practicality of the concept.

\section*{ACKNOWLEDGMENT}

The authors wish to acknowledge Dr. Sylvester Ashok and Dr. Daniel Schrage from the Integrated Product Life-Cycle Engineering (IPLE) Laboratory at Georgia Institute of Technology for providing us with the tools necessary for rapid prototyping of the test vehicle.

\end{document}